\documentclass[11pt,a4paper]{article}
\usepackage[hyperref]{acl}
\usepackage{times}

\usepackage{microtype}
\usepackage{array}
\usepackage{graphicx}
\usepackage{microtype}
\usepackage{caption, subcaption} 
\usepackage{microtype}

\title{Hypers at ComMA@ICON: Modelling Aggressiveness, Gender Bias\\ and Communal Bias Identification }

\author{Sean Benhur\(^1\)\thanks{Equal Contribution}, \hspace{2pt} Roshan Nayak\(^2\)\(^*\), \hspace{2pt} Kanchana Sivanraju\(^1\) \\ \bf Adeep Hande\(^3\), \bf Subalalitha Chinnaudayar Navaneethakrishnan\(^4\),\\ \textbf{Ruba Priyadharshini\(^5\)}, \textbf{Bharathi Raja Chakravarthi\(^6\)} \\ \(^1\)PSG College of Arts and Science \(^2\)BMS College of Engineering\\ \(^3\)Indian Institute of Information Technology Tiruchirappalli \\\(^4\) SRM Institute of Science and Technology, India   \(^5\)ULTRA Arts and Science College \\\(^6\)National University of Ireland Galway \\
  \texttt{seanbenhur@gmail.com} \\}

\begin{document}
\maketitle
\begin{abstract}
Due to the exponentially increasing reach of social media, it is essential to focus on its negative aspects as it can potentially divide society and incite people into violence. In this paper, we present our system description of work on the shared task ComMA@ICON, where we have to classify how aggressive the sentence is and if the sentence is gender-biased or communal-biased.These three could be the primary reasons to cause significant problems in society. As team Hypers we have proposed an approach which utilizes different pretrained models with Attention and mean pooling methods. We were able to get Rank 3 with 0.223 Instance F1 score on Bengali, Rank 2 with 0.322 Instance F1 score on Multi-lingual set, Rank 4 with 0.129 Instance F1 score on Meitei and Rank 5 with 0.336 Instance F1 score on Hindi. The source code and the pretrained models of this work can be found here\footnote{\url{https://github.com/seanbenhur/multilingual_aggresive_gender_bias_communal_bias_identifcation}}.
\end{abstract}

\section{Introduction}
\begin{table*}[]
    \centering
    \begin{tabular}{ | m{15em} | m{2cm} | m{2cm} | m{2cm} | m{2cm} | }
\hline
\textbf{Text} &  \textbf{Language} & \textbf{Aggressive}  & 
\textbf{Gender Bias} & \textbf{Communal Bias}\\
\hline
angakpa nupini eiga unaradi fadoubi & Meitei & CAG & NGEN & NCOM\\
\hline
hehhh ym pkte nupi ng & Meitei & CAG & GEN & NCOM\\
\hline
tome hola bal chera tumi nijai jante & Bengali & OAG & GEN & NCOM\\
\hline
you know to whom im addressing, 'ye hindustan ke liye dimak h jo usko ander se khokla krre h' \#muslimvirus & Hindi & OAG & NGEN & COM\\
\hline
mulle tere allah ki ma ka bhosda & Hindi & OAG & GEN & COM\\
\hline
gudmarani chale ke maaaar & Multi & OAG & GEN & NCOM\\
\hline
jay bheem namo buddhay & Multi & OAG & NGEN & NCOM\\
\hline
\end{tabular}
    \caption{Samples from the dataset and their corresponding class labels for each of the tasks.}
    \label{tab:table1}
\end{table*}

\begin{table*}[]
    \centering
%    \titlebox{Samples distribution in training set.}
    \begin{tabular}{ l l l l l l l l l l }
\hline
\textbf{Language} & \textbf{NAG} & \textbf{CAG} & \textbf{OAG} & \textbf{NGEN} & \textbf{GEN} & \textbf{NCOM} & \textbf{COM} & \textbf{Total}\\
\hline
Meitei & 1,258 & 1,495 & 456 & 3,006 & 203 & 2,967 & 242 & 3,209\\
Bengali & 1,115 & 494 & 1,782 & 2,120 & 1,271 & 2,975 & 416 & 3,391\\
Hindi & 1,594 & 969 & 3,052 & 4,440 & 1,175 & 4,402 & 1,213 & 5,615\\
Multi-lang & 3,966 & 2,956 & 5,289 & 9,564 & 2,647 & 10,342 & 1,869 & 12,211\\
\hline
\end{tabular}
    \caption{Samples distribution in the training set.}
    \label{tab:table2}
\end{table*}

The Internet is a vast network that connects devices all over the world. Due to mobile technology and affordable internet plans, users can access the Internet with ease, leading to the tremendous growth of the Internet, which is unprecedented. As of January 2021, there were 4.66 billion active internet users, 59.5\% of the world's population. Users would undoubtedly want to increase their reach virtually, and hence the interaction among the people would increase. The people these days are more vocal and, at any cost, want their voices or opinions to be reached to a multitude of people. Hence, people search for a platform to share their views, and social media is an ideal place for that. This exact mindset of people has fueled the copious amounts of social media users globally.

Social media are the technologies that allow the creation, sharing, or exchange of information, interests, ideas, and other forms of expression. Its use is an ever-increasing phenomenon of the 21st century \cite{sonia2010rise}. There are a plethora of social media platforms, each attracting people in unique ways. As of January 2021, there were 4.2 billion active social media users. Considering the reach of social media, they can spread people's opinions in a few minutes \cite{boyang2015spread}. Hence it will have a significant effect on society which could be both positive as well as harmful \cite{jyoti2017impact}. But there are instances in which the situation would go out of hand. For example, people could differ in their opinions, and people with similar views tend to form a group to denounce the group with ideas that are not the same as theirs. During the denouncement, there is a possibility that a user could show his improper behaviour, thus making offensive \cite{adeep2021offensive}, misogynistic \cite{elena2019misogyny}, hateful \cite{bhatia2021rule}, or any kind of statements that has the potential to create controversy \cite{mauro2017controversy}. Such statements may be intended towards an individual or a group and are not considered to be good or acceptable in the society. As such behaviour would influence others wrongly and instigate violence or affect the mental health, leading to unpleasant situations. Hence it is necessary to flag such posts and its advisable to take them down from the social media platform and also retribute the user responsible for such posts\cite{hande2021benchmarking}. There could be several reasons that a post by the user is considered inappropriate. Considering how important it is to regulate toxic post, in this paper we will be presenting a system to identify if the user is being aggressive on some individual or a community, or being biased regarding the gender \cite{mochamad2020gender,hande-etal-2020-kancmd}, or is targeting a particular religion or caste \cite{roy2016social}. 

Undoubtedly, English is the widely spoken language in the world \cite{crystal_2008}. But as there are no hardbound rules that users must text in English, the text found on social media could be multilingual and lack grammatical rules \cite{siti2020errors}. Also, there could be unwanted symbols in the text \cite{chakravarthi2021dravidiancodemix}. Considering all such challenges, in this paper, we present a model to classify the multi-lingual sentence written by the user as to how aggressive it is and if it is gendered and communal oriented text. The dataset had multilingual texts with the code-mix of English and several other languages native to India. Meitei and Bangla are native to the Indian states of Manipur and West Bengal, respectively, whereas Hindi is predominant in Northern India.

The rest of the paper is structured as follows, section \ref{section:2} describes about dataset used for the shared task. The section \ref{section:3} describes the models and architectures that were used for the tasks. In section \ref{section:4} we discuss about the results obtained during the study, and the last section \ref{section:5} concludes the work.
\begin{table*}[htbp]
    \centering
   % \titlebox{Samples distribution in validation set.}
    \begin{tabular}{ l l l l l l l l l l }
\hline
\textbf{Language} & \textbf{NAG} & \textbf{CAG} & \textbf{OAG} & \textbf{NGEN} & \textbf{GEN} & \textbf{NCOM} & \textbf{COM} & \textbf{Total}\\
\hline
Meitei & 370 & 471 & 159 & 945 & 55 & 932 & 68 & 1,000\\
Bengali & 333 & 157 & 501 & 624 & 367 & 879 & 112 & 991\\
Hindi & 305 & 167 & 526 & 775 & 225 & 804 & 196 & 998\\
Multi & 1,007 & 797 & 1,193 & 2,349 & 648 & 2,622 & 375 & 2,997\\
\hline
\end{tabular}
    \caption{Distribution of samples in the dev set.}
    \label{tab:table3}
\end{table*}

\begin{table*}[htbp]
    \centering
    %\titlebox{Results of trained models on validation set.}
    \scalebox{0.7}{\begin{tabular}{ l | l | l | l | l | l | l | l | l | l | l | l | l | l | l }
\hline
  & & \textbf{Language} & \multicolumn{3}{c|}{\textbf{AGG}} & \multicolumn{3}{c|}{\textbf{GB}} & \multicolumn{3}{c|}{\textbf{CB}} & \multicolumn{3}{c}{\textbf{Overall}}\\
\hline
\textbf{Model} & \textbf{Arch}  & & P & R & F1 & P & R & F1 & P & R & F1 & P & R & F1\\
\hline
MURIL & AP & Meitei & 0.470 & 0.470 & 0.470 & 0.599 & 0.599 & 0.599 & 0.493 & 0.493 & 0.493 & \textbf{0.521} & \textbf{0.521} & \textbf{0.521}\\
\hline
MURIL & MP & Meitei & 0.471 & 0.471 & 0.471 & 0.603 & 0.603 & 0.603 & 0.356 & 0.356 & 0.356 & 0.477 & 0.477 & 0.477\\
\hline
csebuetnlp/banglabert & AP & Bengali & 0.642 & 0.642 & 0.642 & 0.755 & 0.755 & 0.755 & 0.692 & 0.692 & 0.692 & \textbf{0.696} &  \textbf{0.696} & \textbf{0.696}\\
\hline
csebuetnlp/banglabert & MP & Bengali & 0.635 & 0.635 & 0.635 & 0.762 & 0.762 & 0.762 & 0.612 & 0.612 & 0.612 & 0.670 & 0.670 & 0.670\\
\hline
MURIL & AP & Hindi & 0.594 & 0.594 & 0.594 & 0.816 & 0.816 & 0.816 & 0.909 & 0.909 & 0.909 & 0.773 & 0.773 & 0.773\\
\hline
MURIL & MP & Hindi & 0.683 & 0.683 & 0.683 & 0.827 & 0.827 & 0.827 & 0.902 & 0.902 & 0.902 & \textbf{0.804} & \textbf{0.804} & \textbf{0.804}\\
\hline
MURIL & AP & Multi-Lang &0.618 & 0.618 & 0.618 &0.839 & 0.839& 0.839 &0.661 &0.661 & 0.661 & 0.706&0.706 &0.706\\
\hline
MURIL & MP & Multi-Lang &0.612 &0.612 &0.612 &0.823 & 0.823&0.823 &0.891 &0.891 & 0.891 & \textbf{0.791} & \textbf{0.791} & \textbf{0.791}\\
\hline
\end{tabular}}
    \caption{Results on the dev set. AGG: Aggressive, GB: Gender Bias, CB: Communal Bias, Arch: Architecture, AP: Attention-pooling, MP: Mean-pooling. Metrics, Micro average scores of P: Precision, R: Recall, F1: F1-score calculated on the dev set. Overall scores are the average of the aggressive, gender bias, and communal bias.}
    \label{tab:table4}
\end{table*}
\section{Dataset}
\label{section:2}
The ComMA dataset was provided in this task \cite{comma2021td, b-etal-2021-overview}. The dataset had annotations for aggression, gender bias, and communal bias identification for multi-lingual social media sentences\cite{kumar2021comma}. The dataset comprises of code-mixed sentences has 15,000 code-mixed sentences. It is divided into 12,000 sentences for development and 3,000 sentences for the test. The data is divided into four sets, namely Meitei, Bengali, Hindi, and Multi-lingual. The Multi-lingual set comprises sentences of all three languages. The Table \ref{tab:table1} gives an idea of how the dataset could look like. The sentences in every set are classified into one of the classes for each of the three tasks. The tasks and their classes include,
\begin{itemize}
  \item \textbf{Aggression Classification}: The text is divided into Overtly Aggressive (OAG),  Covertly Aggressive (CAG) or Non-aggressive (NAG)\\
  \item \textbf{Gender Bias Classification}: The text is divided into gendered (GEN) or non-gendered (NGEN).\\
  \item \textbf{Communal Bias Classification}: The text is divided into communal (COM) or non-communal (NCOM)\\
\end{itemize}
The samples count of classes is far from equal. Hence the dataset is quite imbalanced. The dataset distribution is displayed in the Table \ref{tab:table2}.

\section{Methodology}
\label{section:3}
In this section, we describe the methodology of our systems, including data preprocessing and Model architecture. We use mean pooled, and Attention pooled pretrained models, which was shown to provide better results \cite{benhur2021psgdravidiancodemixhasoc2021}. We trained all the models with the batch size of 8, dropout of 0.3, linear scheduler for learning rate scheduling with 2e-5 as an initial learning rate.

\subsection{Data Preprocessing}
The task dataset consists of both codemixed and native scripts; for the Bengali dataset, we converted the emojis into Bengali language using bnemo GitHub repository\footnote{\url{https://github.com/faruk-ahmad/bnemo}}, we removed URLs and punctuations in the text for all the languages. Since the dataset is imbalanced, we sampled the dataset uniformly.

\subsection{Pretrained Models}
We finetuned pretrained transformers with custom poolers on hidden states on MURIL \cite{khanuja2021muril} for Hindi, Meitei and Multilingual datasets and BanglaBert \cite{abhik2021bert} for Bengali dataset. In this section, we describe our Pooling methods and pretrained models.

\subsubsection{MURIL}
MURIL is a pretrained model, specifically made for Indian languages. MuRIL, the pretrained model, is trained in 16 different Indian Languages. Instead of the usual Masked Language Modelling approach, the model is trained on both Masked Language Modelling(MLM) objective and Translated Language Modelling(TLM) objective. In the TLM objective, both translated and transliterated sentence pairs are sent to the model for training. This model outperforms mBERT on all the tasks for Indian languages on the XTREME \cite{hu2020xtreme} benchmark.

\begin{table*}[htbp]
    \centering
    %\titlebox{Results of trained models on validation set.}
    \scalebox{0.8}{\begin{tabular}{ l | l | l | l | l }
\hline
  \textbf{Model} & \textbf{Architecture} & \textbf{Language} & \textbf{Overall Micro F1} & \textbf{Overall Instance F1}\\
\hline
MURIL & Attention-pooler & Meitei & \textbf{0.472} & \textbf{0.129}\\
\hline
MURIL & Mean-pooler & Meitei & 0.436 & 0.080\\
\hline
csebuetnlp/banglabert & Attention-pooler & Bengali &  \textbf{0.579} & \textbf{0.223}\\
\hline
csebuetnlp/banglabert & Mean-pooler & Bengali & 0.572 & 0.201\\
\hline
MURIL & Attention-pooler & Hindi & 0.662 & 0.326\\
\hline
MURIL & Mean-pooler & Hindi & \textbf{0.683} & \textbf{0.336}\\
\hline
MURIL & Attention-pooler & Multi-Lang & \textbf{0.685} & \textbf{0.322}\\
\hline
MURIL & Mean-pooler & Multi-Lang & 0.601 & 0.280\\
\hline
\end{tabular}}
    \caption{Results on the test set. Overall, the Micro F1 score is calculated by the average of aggressive, gender, and communal biases. Instance F1 score is similar to the F1 score but when all the three labels are  predicted correctly.}
    \label{tab:table5}
\end{table*}

\begin{table*}[htbp]
    \centering
    %\titlebox{Results of trained models on validation set.}
    \scalebox{1.0}{\begin{tabular}{ l | l | l }
\hline
  \textbf{Language} & \textbf{Overall Micro F1} & \textbf{Overall Instance F1}\\
  \hline
   Meitei & 0.472 & 0.129\\
  \hline
  Bangla & 0.579 & 0.223\\
  \hline
  Hindi & 0.683 & 0.336\\
  \hline
  Multi-Lang & 0.685 & 0.322\\
  \hline
\end{tabular}}
    \caption{Results obtained when submitted to the competition.}
    \label{tab:table6}
\end{table*}

\subsubsection{BanglaBert}
Banglabert is pretrained on more than 18.5 GB of a corpus in Bengali corpora. Banglabert achieves the state of the art performance on Bengali texts on five downstream tasks. It outperforms multilingual baselines with more than a 3.5 percentage score.
Banglabert is pretrained using ELECTRA \cite{clark2020electra} with Replaced Token Detection(RTD) objective. In this setup, two networks, a generator network and discriminator network, are used, while training both the networks are trained jointly. The generator is trained on the Masked Language Modelling objective, where a portion of the tokens in the sentence is masked and is asked to predict the masked tokens using the rest of the input. The masked
tickets are replaced by tokens sampled from the
generator's output distribution for the corresponding masks for the discriminator input. The discriminator then has to predict
whether each token is from the original sequence
or not. After pretraining, the discriminator is used for finetuning. 

\subsection{Attention Pooler}
The attention operation described in equation 1 is applied to the CLS token in last hidden state of the pretrained transformer; we hypothesize that this helps the model learn the contribution of individual tokens. Finally, the returned pooled output from the transformer is further passed to a linear layer to predict the label.
\begin{equation}
   o = W^T_{h} softmax(qh^T_{CLS})h_{CLS}
\end{equation}
where $W^T_{h}$ and $q$ are learnable weights and $h_{CLS}$ is the CLS representation and $o$ is the output.
\begin{equation}
    y = softmax(W^T_{o} + bo)
\end{equation}

\subsection{Mean Pooler}
In the mean-pooling method, the last hidden state of the tokens are averaged on each sentence, and it is passed onto the linear layer to output the final probabilities.

\section{Results}
\label{section:4}
Pretrained models with different pooling methods were trained on each language set and then validated on dev sets. For the competition submissions, we submitted the model with a higher Micro F1 score on the dev set. Table \ref{tab:table4} shows the results of the dev set, and the Table \ref{tab:table3} depicts the data distribution of the set used to validate the trained models. The training process was done on Tesla P100 GPU. In the test set submissions, We were able to get Rank 3 with 0.223 InstanceF1 score on Bengali,  Rank 2 with 0.322 Instance F1 score on Multi-lingual set,  Rank 4 with  0.129  Instance  F1  score  on  Meitei  and Rank 5 with 0.336 Instance F1 score on Hindi. The competition results are shown in Table \ref{tab:table6}. Table \ref{tab:table5} shows the Overall Micro F1 score and Instance F1 score on the test set. The pretrained model MURIL was not trained on Meitei, but it still achieved comparable performance on the Test set; we hypothesize that since MURIL was trained both on transliterated pairs on TLM objective and the Meitei dataset also only consisted of code-mixed texts, we get a fair results on meitei test set.

\section{Conclusion}
In this paper, we experimented with different pooling methods, namely Attention Pooling and Mean Pooling and pretrained models, to classify sentences, how aggressive they are, and whether gender-oriented or communal. From Table \ref{tab:table5} its evident that attention pooling worked better in most of the cases. We have also discussed the various essential reasons why the work on this is necessary. As for future work, we will consider improving our scores, especially on multilingual and meitei datasets, and experimenting with other pretrained models.

\bibliography{anthology,custom}
\bibliographystyle{acl_natbib}

\end{document}